\documentclass{llncs}
\usepackage{makeidx}

\usepackage{times}

\usepackage{balance}  
\usepackage{graphicx} 
\usepackage{times}    
\usepackage{multirow}
\usepackage{url}      
\usepackage{rotating}

\makeatletter
\def\url@leostyle{%
  \@ifundefined{selectfont}{\def\UrlFont{\sf}}{\def\UrlFont{\small\bf\ttfamily}}}
\makeatother
\usepackage{algorithm}
\usepackage[noend]{algpseudocode}

\newcommand{\ignore}[1]{}
\newcommand{\bi}{\begin{itemize}}
\newcommand{\ei}{\end{itemize}}
\newcommand{\be}{\begin{enumerate}}
\newcommand{\ee}{\end{enumerate}}

\ignore{
\usepackage[pdftex]{hyperref}
\hypersetup{
pdftitle={SIGCHI Conference Proceedings Format},
pdfauthor={LaTeX},
pdfkeywords={SIGCHI, proceedings, archival format},
bookmarksnumbered,
pdfstartview={FitH},
colorlinks,
citecolor=black,
filecolor=black,
linkcolor=black,
urlcolor=black,
breaklinks=true,
}
}

\begin{document}

\title{Towards A Virtual Assistant That Can Be Taught \\
New Tasks In Any Domain By Its End-Users\thanks{Many thanks to the participants of our usability study.  Also thanks
to Patrick Haffner, Michael Johnston, Hyuckchul Jung, Amanda Stent,
and Svetlana Stoyanchev for helpful discussions.}
}
\author{
I. Dan Melamed\inst{1,3} and Nobal B. Niraula\inst{2,3}
}
\institute{
Microsoft Research, New York, NY, U.S.A. \\
\email{\{lastname\}@microsoft.com}
\and
University of Memphis, TN, U.S.A. \\
\email{nbnraula@memphis.edu}
\and
Much of this work was done while both authors were working at AT\&T Labs-Research.
}

\maketitle

\begin{abstract}
The challenge stated in the title can be divided into two main
problems.  The first problem is to reliably mimic the way that users
interact with user interfaces. The second problem is to build an
instructible agent, i.e.\ one that can be taught to execute tasks
expressed as previously unseen natural language commands.  This paper
proposes a solution to the second problem, a system we call Helpa.
End-users can teach Helpa arbitrary new tasks whose level of
complexity is similar to the tasks available from today's most popular
virtual assistants.  Teaching Helpa does not involve any programming.
Instead, users teach Helpa by providing just one example of
a command paired with a demonstration of how to execute that command.
Helpa does not rely on any pre-existing domain-specific knowledge. It
is therefore completely domain-independent. Our usability study showed
that end-users can teach Helpa many new tasks in less than a minute
each, often much less.
\end{abstract}

\section{Introduction}
\label{intro}


Popular virtual assistants (VAs), such as
Siri\footnote{\url{www.apple.com/ios/siri/}},
Cortana\footnote{\url{windows.microsoft.com/en-us/windows-10/}
\newline \url{getstarted-what-is-cortana}}, and
GoogleNow\footnote{\url{www.google.com/landing/now}}, can perform
dozens of different tasks, such as finding directions and making
restaurant reservations.  These are the tasks that the VA developers
expected to be the most widely used.  However, every VA user can
probably think of one or more other tasks that they would like their
VA to help with, which the developers have simply not implemented yet.
The unavailable tasks are as varied as the users.  Thus, the demand
curve for VA tasks has a very long and heavy tail of unsatisfied
demand.  The capabilities of currently available VAs represent only a
tiny fraction of their potential. Even the available tasks are often
implemented differently from how users would prefer.

This situation is unavoidable given how VAs are currently developed.
There will never be enough VA developers to customize VAs in all the
ways that users would like.  The only way to close the gap between
what VAs can do and what users want them to do is to enable
non-technical end-users to teach new tasks to their VAs.  Many users
would be willing and able to do so, if it were as quick and easy as
teaching a person.

The most common way to teach a person a relatively simple new task is
to describe the task and then demonstrate how to do it.  For decades,
researchers have been trying to build computer systems that can be
taught the same way.  Their efforts comprise a body of work most
commonly referred to as ``programming by demonstration''
(PBD)\cite{Curry78}. \footnote{``PBD'' is an unfortunate name, because
  most of the non-technical users that can benefit from it are
  reluctant to attempt anything with ``programming'' in its name.}

\begin{table*}[tb]
\centering
\begin{tabular}{r|c|c|c|} 
\bf{program class $\rightarrow$} & \bf{variable-free} &\bf{non-branching} & \bf{branching}
  \\ \hline
\bf{finite set of tasks} & & Siri, Cortana, \it{et al.} & \\ \hline
\bf{domain-restricted set of tasks} & & & PLOW \it{et al.} \\ \hline
\bf{most tasks} & & Helpa & \\ \hline
\bf{all tasks} & macros & & \\ \hline
\end{tabular}
\caption{
\label{expPower}
Virtual assistants trade off task expressive power for task domain-dependence.}
\vspace*{-5mm} 
\end{table*}

The simplest kind of PBD system creates and runs programs with no
variables, colloquially known as macros. The absence of variables,
which also implies the absence of loops and conditionals, makes it
easier for non-programmers to understand and use macros.
Nevertheless, macros see little use outside of special environments
such as text editing software, because there are relatively few
situations in which a program without variables can be useful.

To increase the usefulness of PBD, researchers have attempted to build
systems that can be taught more powerful classes of programs, all the
way up to Turing-equivalent systems with variables, loops, and
conditionals (e.g., see \cite{CypherHalbert93,Cypher+2010} and
references therein).  Invariably, such attempts run into the
limitations of the current state of the art in natural language
understanding.  At present, the only known way for computers to deal
with the richness of language that people use to describe complex
tasks is to limit the tasks to a narrow domain, such as
travel reservations or messaging. For example, the PLOW system
\cite{Allen+07} is powerful enough to learn programs with variables,
loops, and subroutines.  Yet, it can learn tasks only within the task
domains covered by its ontology.  In order to demonstrate PLOW's
ability to learn tasks in a new domain, its authors had to manually
extend its ontology to the new domain.  To the best of our knowledge,
all previous PBD systems with variables are similarly limited to at
most a handful of task domains.\footnote{From the point of view of
most users, who do not have access to the developers.}

The class of variable-free programs and the class of Turing-equivalent
programs are the two extremes on a continuum of expressive power.
However, most of the tasks available from today's most popular VAs can
be expressed by programs that are in another class between those two
extremes.  These programs are in the ``non-branching'' class, where
programs can have variables but cannot have loops or conditionals.
Judging by the popularity of VA software, a very large number of
people could benefit from a VA that can be taught new non-branching
programs by its end-users.

This paper presents Helpa, a system that can be taught non-branching
programs via PBD.  We have developed a way to teach such programs
without any prior domain knowledge, which works surprisingly well in
most cases. Therefore, Helpa imposes no restrictions on the domains in
which users can teach it new tasks. We believe that Helpa's innovative
trade-off of expressive power for domain-independence occupies a sweet
spot of very high utility, compared to the other classes of VAs in
Table~\ref{expPower}. In addition, our usability study showed that
Helpa's design makes it possible for end-users to teach it many new
tasks in less than a minute each --- fast enough for practical use in
the real world.

Following \cite{HuffmanLaird92}, we shall refer to the teachable
component of a VA as an instructible agent (IA), and the challenge of
building an IA as the IA problem.  After formalizing this problem in
the next section, we shall describe our proposed solution. We shall
then describe some of its current limitations, which explain why we
claim that Helpa can learn only ``most tasks'', rather than ``all
tasks'', in Table~\ref{expPower}. Lastly, we shall describe a
usability study that we carried out to evaluate Helpa's effectiveness.

\section{The Instructible Agent (IA) Problem}
\label{sec:iap}

The IA problem is to build a system that can correctly execute a task
expressed as a previously unseen natural language command.  We shall
put aside the question of what counts as natural language by accepting
any string of symbols as a command.  It is more challenging to
operationalize the notion of executing a task.

Every PBD system interacts with a particular user interface (UI).  It
records the user's actions in that UI when a user is demonstrating a
new task for it to learn.  It mimics the user's actions in that UI to
execute tasks that it has learned.  Reliably interacting with a UI in
this manner is a challenging problem (e.g., see \cite{Niraula+14}).
The present work makes no attempt to solve it.
Rather, we abstract the notion of task execution into a data structure
that we call a ``UI script''.  We assume that when a PBD system
records a user's actions, the result is a UI script.  And when it's
time for a PBD system to mimic a user's actions back to the UI, it
does so by reading and executing a UI script.  

Since all of the IA's interactions with the UI are via a UI script, we
can define the IA problem independently of the problem of reliably
interacting with the UI.  In particular, we define the IA problem as
predicting a UI script from a command.  \cite{Branavan+09} studied a
special case of this problem where the natural language input
explicitly referred to every user action in the UI script.
\cite{LauDomingos00} and others have studied a related but different
problem where the goal was to predict programs from program traces.
IAs that aim to learn branching programs must predict
branching UI scripts but, in the present work, we limit our attention
to non-branching programs and non-branching UI scripts.

\section{Helpa}
\label{helpa}

\subsection{Model}
\label{defs}

Given sufficient training data, it might be possible to solve the IA
problem via machine techniques (e.g., \cite{Bengio+13}).  We are not
aware of any pre-existing training data 
for this problem.  To compensate for the lack of data, we used a model
with very strong biases, so that it can be learned from only one
example (per task) of the kind that we might reasonably expect a non-technical
end-user to provide.  The Helpa model has three parts for every task
$t$: \be
\item The class ${\cal T}_t$ of commands that pertain to $t$.  We
  shall encode ${\cal T}_t$ in a data structure called a ``command
  template''.
\item The class ${\cal P}_t$ of UI scripts for $t$.  We shall
  encode ${\cal P}_t$ as a non-branching program.
\item A mapping of variables between ${\cal T}_t$ and ${\cal P}_t$,
  which we call a ``variable binding function.''
\ee
We shall now expand on each of these concepts.

A natural language command given to an IA can be segmented into
constants and variable values.  Variable values are words or phrases
that are likely to vary among commands from
the same class.  Constants are ``filler'' language that is likely to
remain the same for every command in the class.  For example, suppose
a user wants to train her system to check flight arrival times using
the command ``When does KLM flight 213 land?'' In this command, ``KLM''
and ``213'' are variable values.  The other symbols are constants.
A {\bf command template} can be derived from a command by replacing
each variable value with the name of a variable.  ``When does
$X_1$ flight $X_2$ land?'' is a command template for the previous
example.

To justify our use of the term ``program'', we must first say more
about UI scripts. In the present work, we limit our attention to UIs
that consist of discrete elements, where all user actions are
unambiguously separate from each other and happen one at a
time\footnote{A smart-phone touchscreen or a web browser would fit this
  description, for example, but a motion-capture suit would not.}.  A
non-branching {\bf UI script} for such a UI is a sequence of actions, where every
action pertains to at most one element of the UI.  E.g., a UI script
for a web browser might involve an action pertaining to the 4th text
field currently displayed and an action pertaining to the leftmost
pull-down menu.  A common action that does not pertain to a
specific UI element is to wait for some condition to occur in the UI,
such as waiting for a web page to load.  Besides identifying an
element in the UI, each action can also specify a parameter value,
such as what to type into the text field or how long to wait for the
page to load\footnote{More generally, each action can have multiple
  parameter values.  We omit this generalization for simplicity of
  exposition.}.  An example of a UI script is in Figure~\ref{demo}.
\begin{figure}[tb]
\centering 
\begin{tabular}{rcl}
action type & UI element & parameter value \\
\hline
textbox\_fill & address\_bar & \url{flightarrivals.com} \\
wait\_for & & page\_load \\
select\_from & menu\_1 & KLM \\
textbox\_fill & textbox\_1 & 213 \\
click\_button & button\_1 & \\
wait\_for & & page\_load \\
\hline
\end{tabular}
\caption{
\label{demo}
Example of a UI script for the command ``When does KLM
  flight 213 land?''}
\end{figure}

Every non-branching program is also just a sequence of actions.  A
program differs from a UI script only in that some of the parameter
values can be variables. E.g., to create a program from the UI
script in Figure~\ref{demo}, we would replace the parameter value
``KLM'' with a variable name like $X_1$ and the parameter value
``213'' with another variable name like $X_2$. Replacing values with
variable names, both in commands and in UI scripts, is a form of
generalization.  This kind of generalization is the most common way
for PBD systems to learn (e.g., \cite{Nix83}).

Finally, a {\bf variable binding function} maps the variables in a
command template to the variables in a program.  Helpa allows a
command template variable to map to multiple program variables, but
not vice versa.  The one-to-many mapping can be useful, e.g., when a
web form asks for a shipping address separately from a billing
address, and the user always wants to use the same address for both.
We do not allow multiple command template variables to map to the same
program variable. Doing so would merely increase system complexity
without any benefits.

\subsection{System Architecture and Components}
\label{system}

\begin{figure}[tb]
\centering 
\includegraphics[width=3.25in]{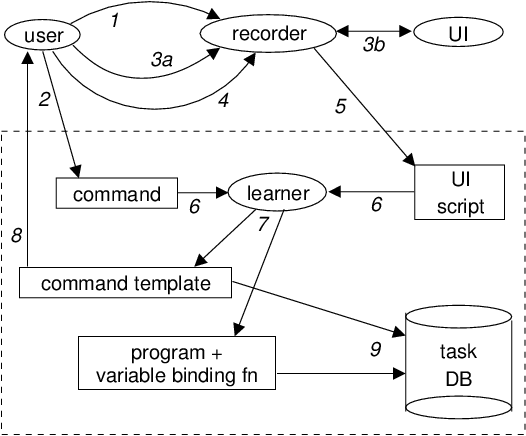}
\caption{
\label{dfd-train}
Data flow diagram for Helpa's learning mode.
}
\end{figure}

\begin{figure}[tb]
\centering 
\includegraphics[width=3.25in]{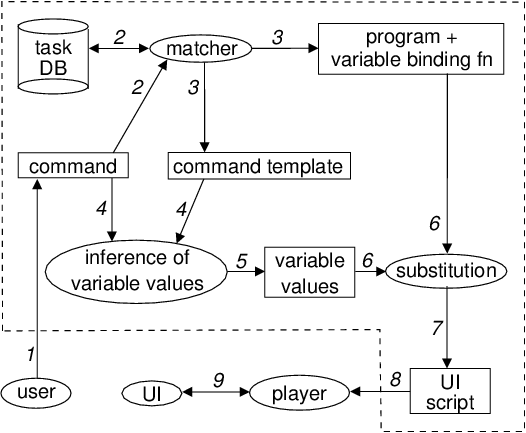}
\caption{
\label{dfd-run}
Data flow diagram for Helpa's execution mode.
}
\end{figure}

With the Helpa model in mind, we can describe how Helpa works.  It has
two modes of operation: learning and execution, illustrated in
Figures~\ref{dfd-train} and~\ref{dfd-run}, respectively.  In both
figures, dashed lines delimit the Helpa system boundary, and numbers
indicate the order of events.  Both modes use a database of tasks,
where every record consists of a command template, a program, and a
variable binding function.  Tasks are created in learning mode and
executed in execution mode.  

The user initiates the learning mode by starting the UI recorder (1).
The user then provides an example command (2) and demonstrates how to
execute the command (3a).  During the demo, the recorder is
transparent to the user and to the UI.  It records all user actions
and any relevant responses from the UI (3b).  When the user stops the
recorder (4), the recorder writes a UI script (5).  Then, the learner
takes the example command and the UI script (6), and infers a command
template, a program, and a variable binding function for the task (7).
The command template is shown to the user for approval (8).  If the
user approves, then the program and variable binding function are
stored in the task database, keyed on the command template (9).
Otherwise, the user can start over.

Execution mode
starts when the user provides a new command (1) without starting the
UI recorder.  The matcher queries the task database (2) and selects
the task whose command template matches the new command (3).  The
command template for that task is compared to the new command (4), in
order to infer the variable values (5).  Currently, the values are
inferred merely by deleting the constant parts of the command template
from the command.  Once found, the values are substituted into the
program via the variable binding function (6) to create a new UI
script (7).  The UI script is sent to the player (8), which mimics the
way that a user would execute that task in the UI (9).  Thus, after
learning a new task, and storing it keyed on its command template,
Helpa can execute new commands matching that template, with previously
unseen parameter values.

We shall now say more about some of the subsystems that our diagrams
refer to.  The diagrams show the player and recorder outside of the
Helpa system boundary, because we do not consider these components to
be part of Helpa.  A different player and recorder are necessary for
every type of UI.  However, regardless of the UI, Helpa interacts with
the world only through UI scripts.  Therefore, Helpa is
UI-independent, which also makes it device-independent.

In execution mode, the matcher looks for a command template that can
be made identical to the command by substituting the template's
variables with some of the command's substrings.  E.g., the template
``When does $X_1$ flight $X_2$ land?'' can be made identical to the
command ``When does United flight 555 land?'' by substituting $X_1$
with ``United'' and $X_2$ with ``555''.  This kind of matching is a
special case of unification, for which efficient algorithms exist
\cite{Robinson71}.

Figure~\ref{dfd-run} shows only what happens if
exactly one unifying template is found.  Otherwise, control passes to
a clarification subsystem, which is not shown in the diagram.  If no
suitable template is found, this subsystem provides a list of
available command templates to the user, in order of string similarity
to the command, and offers the user a chance to try another command.
If multiple templates unify with the new command, they are displayed
in order of their amount of overlapping filler text, and the user is
asked to disambiguate their command by rewording it.

The learner used in learning mode is responsible for generalizing the
command to a command template, generalizing the UI script to a
program, and deciding which variables in the command template
correspond to which variables in the program.  A key insight that
makes it possible to learn from only one example is that, typically,
each variable value in the example command is the same as a parameter
value in the UI script.  In contrast, the constant parts of the
command typically bear no resemblance to the rest of the UI script.

\begin{algorithm}[tb]
\begin{algorithmic}[1]
\Require command $C$, UI script $S$
\State $L_1 = L_2 = \emptyset$ \Comment{empty lists}
\For {$i = 1$ to $|D|$}
\State $q \gets$ value of parameter in action $i$ of $S$
\If {$q$ matches $C$ from word $m$ to word $n$}
\State $len \gets m - n + 1$
\State $L_1$.append$(\langle len, i, m, n \rangle)$ \Comment{list of 4-tuples}
\EndIf
\EndFor
\State sort $L_1$ on $len$
\State $R[1.. |C|] \gets \vec{0}$  \Comment{array of $|C|$ zeros}
\ForAll { $\langle len, i, m, n \rangle \in L_1$ }
\If {\parbox[t]{\linewidth}{
$R[m .. n] = \vec{0}$ \\
or \parbox[t]{\linewidth}{
    $\exists d: (R[m .. n] = \vec{d}$ \\
\hspace*{.3in} and $R[m-1] \neq d$ and $R[n+1] \neq d)$}}
\parbox[t]{.2in}}
\State $R[m .. n] \gets \vec{i}$ \Comment{put $i$ in positions $m$
  thru $n$}
\State $L_2$.append$(\langle m, n, i \rangle)$ \Comment{list of triplets}
\EndIf
\EndFor
\State sort $L_2$ on $m$
\State $T = C$ \Comment{command template}
\State $P = D$ \Comment{program}
\State $B = \emptyset$ \Comment{variable binding function}
\ForAll { $\langle m, n, i \rangle \in L_2$ }
\State replace words $m$ thru $n$ of $T$ with ``$X_m$''
\State replace parameter in line $i$ of $P$ with ``$X_m$''
\State add $(``X_m" \rightarrow i)$ to $B$
\EndFor
\Ensure command template $T$, program $P$, variable binding function $B$
\end{algorithmic}
\caption{
\label{algoLearn}
Helpa learning algorithm
}
\end{algorithm}
Helpa's learner uses this insight as shown in
Algorithm~\ref{algoLearn}.  The first loop (lines 2--6) matches the
parameter values in the UI script with substrings of the command, and
stores them in list $L_1$.  After the loop, the list is sorted on the
length of the matching substring, in order to give preference to
longer matches.  The second loop (lines 9--12) traverses $L_1$ in
order from longest match to shortest.  Each matching action attempts
to reserve its substring of the command by filling the corresponding
span of the reservation array $R$ with its action index $i$.  The
reservation attempt succeeds if one of two conditions holds: either
that span is not yet reserved by any other action, or {\em exactly}
that span is reserved by another action (i.e.\ with the same span
boundaries). The latter condition enables one command variable to map
to multiple UI script variables, but only if it's exactly the same
command variable.  Overlapping or nested command variables are not
allowed.  The successful reservations are stored in list $L_2$.  In
line 13, $L_2$ is sorted on the left boundary $m$ of the span of the
variable value in the command.  This order is necessary because, in
execution mode, the variable substitution process assumes that the
order of variables in the variable binding function is the same as the
order of variables in the command template.  The last loop (lines
17-20) traverses $L_2$, whose every element is a mapping from a span
of the command to a line of the UI script.  The learner creates
variable names $X_m$, where $m$ refers to the left boundary of a span
of a command variable.  The learner uses these variable names to
create a command template out of the input command and a program out
of the input UI script.  Naming the variables in this manner allows
one command variable to map to multiple UI script variables.  Since
line 10 disallowed overlapping or nested command variables, there can
be no ambiguity about which command variable each $X_m$ refers to. The
last step in the last loop adds each mapping to the variable binding
function.

\vspace*{-2mm}
\subsection{Limitations}
\label{limits}
At the present stage of development, Helpa has some significant
limitations.  Perhaps the most striking limitation, from a user's
point of view, is that Helpa knows nothing about paraphrasing.  Helpa
doesn't even know that ``April 4, 2016'' is the same as ``04/04/16''.
Likewise,
knowing how to execute ``Find
X'' doesn't help Helpa to execute ``Search for X''.
In order for the learner to work, the variable values in the command
must be identical to the values in the UI script.
\footnote{This
  limitation is not so severe when Helpa is executing a task for the
  same user who trained it on that task, because that user will often
  remember the phrasing that they used.}  The literature offers a
variety of techniques for overcoming this limitation.
For example, we could use
statistical paraphrase generation \cite{Zhao+09} to proactively
expand a newly inferred command template into a set of possible
paraphrases, and store them all in the task database linked to the
same task.  However, the usability study in the next section was
done without the benefit of such techniques.

A more subtle limitation is due to Helpa's simplistic method for
deducing variable values at execution time.  The ``string difference''
method fails when two variables are adjacent in the command template,
because Helpa doesn't know how to partition the adjacent values.
E.g., in a command like ``I need a Ford Taurus Tuesday,'' Helpa has no
way to determine whether ``Taurus'' should be part of the value for
the car variable or part of the value for the day variable.  Again,
there are various natural language processing (NLP) techniques that can
solve most of this problem (e.g., \cite{LingWeld12}). For now, Helpa
works only for commands that have no adjacent variables.  

Although it's easy to think of commands that violate this
constraint, they are relatively rare in practice, at least
in English.  We found long lists of English commands for
Siri\footnote{\url{www.reddit.com/r/iphone/comments/1n43y3/everything_you_can_ask_siri_in_ios_7_fixed}},
for Cortana\footnote{\url{techranker.net/cortana-commands-list-}
\newline  \url{microsoft-voice-commands-video}}, and for
GoogleNow\footnote{\url{forum.xda-developers.com/showthread.php?t=1961636}}.
Two variables were adjacent in only 5 out of 236 Siri commands, in
only 3 out of 91 Cortana commands, and in only 1 out of 98 GoogleNow
commands.  

\section{Usability Study}
\label{eval}

Our working hypothesis in building Helpa was that, in the vast
majority of cases, learning to predict non-branching programs from
natural language commands requires no domain knowledge and only the
most rudimentary NLP.  Our usability study was designed to test this
hypothesis, in terms of Helpa's task completion rates for users who
were not involved in Helpa's development.  We also wanted to measure
how long it takes users to teach new tasks to Helpa.

\subsection{Design of the Study}
\label{exp}

Helpa is UI-independent, but using it with a particular UI
requires a player and recorder for that UI.  A system like Helpa is
most compelling for a speech UI on a mobile device and/or in a
situation where the user's hands are busy.  Unfortunately, we did not
have access to a suitable UI player/recorder for any such UI/device,
and we did not have the resources to create one. The closest
approximation available to us was the Browser Recorder and Player
(BRAP) package.\footnote{\url{https://github.com/nobal/BRAP}}
BRAP records
user actions in a web browser by injecting jQuery code and listening
for JavaScript events such as key-up, select-one, and submit.  This
approach is sufficient for simple web pages, but it often fails on
websites that do not raise events in response to user inputs.
Since BRAP was designed for a
slightly different purpose, it can recognize events related
to only the following HTML elements: text boxes, check boxes, radio
buttons, pull-down menus, and submit buttons.  BRAP knows nothing
about hyperlinks, maps, sliders, calendars, pop-ups, etc.  Even though
BRAP is the most functional software of its kind, its limitations
prevent it from correctly recording demos on most modern websites.

Since BRAP works only with web browsers, our entire study was done in
a Google Chrome web browser, on an Apple MacBook Air computer, through
a keyboard and touchpad.  Also, due to BRAP's limitations, we were
forced to limit our study to websites that used only simple HTML web
forms.  So, we could not use a random sample of web sites, or allow
our study subjects to choose them.

\begin{table*}[tb]
\centering
\begin{tabular}{rcllc}
\hline
 & site type & URL & scenario & \#elts\\
\hline \hline
\multirow{3}{*}{1} & \multirow{3}{0.6in}{\centering mortgage calculator} & \multirow{3}{*}{\parbox{1.75in}{\url{calculator.com/pantaserv/ mortgage.calc}}} & \multirow{3}{*}{\parbox{1.75in}{You are a real estate agent, checking whether your customers can afford certain properties.}} & \multirow{3}{*}{11} \\
& & \\
& & \\
\hline
\multirow{2}{*}{2} & \multirow{2}{*}{thesaurus} & \multirow{2}{*}{\parbox{1.75in}{\url{collinsdictionary.com/english-thesaurus}}} & \multirow{2}{*}{\parbox{1.75in}{You are a writer looking for alternative ways to express yourself.}} & \multirow{2}{*}{4} \\
& & \\
\hline
\multirow{2}{*}{3} & \multirow{2}{*}{book store} & \multirow{2}{*}{\parbox{1.75in}{\url{abebooks.com/servlet/ SearchEntry}}} & \multirow{2}{*}{\parbox{1.75in}{You are a book dealer serving many kinds of readers.}} & \multirow{2}{*}{26} \\
& & \\
\hline
\multirow{3}{*}{4} & \multirow{3}{0.6in}{\centering recruiting} & \multirow{3}{1.75in}{\url{indeed.com/resumes/ advanced}} & \multirow{3}{*}{\parbox{1.75in}{You work for a recruiting firm, searching for candidates to fill various job openings.}} & \multirow{3}{*}{14} \\
& & \\
& & \\
\hline
\multirow{3}{*}{5} & \multirow{3}{0.6in}{\centering investment research} & \multirow{3}{*}{\parbox{1.75in}{\url{nasdaq.com}}} &  \multirow{3}{*}{\parbox{1.75in}{You are an investor who likes to frequently check the prices of your stocks.}} & \multirow{3}{*}{2} \\
& & \\
& & \\
\hline
\multirow{2}{*}{6} & \multirow{2}{0.6in}{\centering scientific database} & \multirow{2}{*}{\parbox{1.75in}{\url{citeseerx.ist.psu.edu/ advanced_search}}} & \multirow{2}{*}{\parbox{1.75in}{You are doing a literature search for a research project.}} & \multirow{2}{*}{13} \\
& & \\
\hline
\multirow{2}{*}{7} & \multirow{2}{*}{car rental} & \multirow{2}{*}{\parbox{1.75in}{\url{priceline.com/l/rental/cars.htm}}} &  \multirow{2}{*}{\parbox{1.75in}{You are a travel agent, researching rental cars.}} & \multirow{2}{*}{7} \\
& & \\
\hline
\multirow{3}{*}{8} & \multirow{3}{0.6in}{\centering cooking recipes} & \multirow{3}{*}{\parbox{1.75in}{\url{allrecipes.com/Search/ Default.aspx?qt=a}}} & \multirow{3}{*}{\parbox{1.75in}{You are in charge of selecting new dishes to put on a restaurant's menu.}} & \multirow{3}{*}{26} \\
& & \\
& & \\
\hline
\multirow{3}{*}{9} & \multirow{3}{*}{airline} & \multirow{3}{1.75in}{\url{united.com/web/ en-US/apps/} \url{booking/flight/ searchOW.aspx}} & \multirow{3}{*}{\parbox{1.75in}{You are a travel agent, checking availability of one-way flights for customers.}} & \multirow{3}{*}{40} \\
& & \\
& & \\
\hline
\multirow{2}{*}{10} & \multirow{2}{*}{dept. store} &
\multirow{2}{*}{\parbox{1.75in}{\url{jcpenney.com}}} &
\multirow{2}{*}{\parbox{1.75in}{You are shopping for gifts for your friends.}} & \multirow{2}{*}{2} \\
& & \\
\hline
\end{tabular}
\caption{
\label{sitelist}
Web sites and scenarios used in our study. \#elts $=$ number of BRAP-compatible UI elements on the landing page.}
\end{table*}

After searching for many hours, we found a sufficiently simple website
in each of 10 diverse categories. For each of these 10 websites, we
picked a scenario for which an IA with variables might be useful.
Table~\ref{sitelist} lists the types of sites we used, along with the
URL, the scenario we picked for each site, and the number of
BRAP-compatible UI elements on the first web page that the study
subjects saw. This study design limited each task to use only one
website, even though Helpa has no such limitation.  Nothing in the
Helpa system was tailored to these websites, these scenarios, or this
study.

We recruited 10 study subjects, and gave them the instructions in
Appendix A.  These instructions were designed to help them get around
Helpa's and BRAP's counter-intuitive limitations.  To summarize,
subjects were instructed that
\bi
\item variable values must appear in the example command exactly the
  same way as they appear in the web form;
\item variables in commands cannot be adjacent; and
\item task demos must use only the HTML elements that BRAP can record.
\item subjects must ignore default values that appear
in web forms.\footnote{BRAP can read default values in web forms, but
  we have not yet figured out a way to determine, without explicit
  indication from the user, whether a given default should become a
  program variable.}  
\ei
We could not think of a way to explain these limitations without
referring to programming concepts.
For this reason, we recruited study subjects from among our
colleagues, all of whom were experienced programmers.

Each subject began by reading the instructions, and asking any
questions they had.  Then, an automated script initialized the task
database to empty, randomized the order of the websites, and guided
the subject through the following protocol for each website:
\be
\item Subject reads the scenario description (Column 4 in
  Table~\ref{sitelist}), and familiarizes themselves with the website.
\item Subject thinks of a task that is relevant to that scenario, and
  of a natural language command that is suitable for that task.
\item Subject interacts with Helpa's learning mode.
\item If the subject disapproves of the command
template that Helpa generated, return to step 1.
\item Subject thinks of another command from the same class.
\item Subject interacts with Helpa's execution mode.
\item Subject provides their opinion on whether Helpa executed the new
  command correctly.
\ee 
The script recorded and timestamped all of the interactions between
Helpa, the study subjects, and the UI.

\subsection{Results}
\label{sec:results}

\setlength{\tabcolsep}{8pt}
\begin{table}[tb]
\centering
\begin{tabular}{|rc|l|llll|ll|}
\hline
&
site type & A & B & C & D & E & F & G 
\\
\hline \hline
1 &
mortgage calculator & 0.5 & 9.5 & 2   & 3.5 & 20   & 90   & 25       
\\
\hline
2 &
thesaurus            & 0.7 & 4   & 3   & 1   & 5.5  & 30   & 26       
\\
\hline
3 &
book store           & 0.9 & 5   & 3   & 2.5 & 9.5  & 31.5 & 35.5     
\\
\hline
4 &
recruiting           & 0.9 & 7   & 3   & 3   & 13.5 & 48.5 & 36       
\\
\hline
5 &
investment research  & 1.0 & 4   & 2   & 1   & 6    & 38   & 37       
\\
\hline
6 &
scientific database  & 1.0 & 6.5 & 2   & 2.5 & 7.5  & 46   & 46.5     
\\
\hline
7 &
car rental           & 1.0 & 8   & 2   & 4   & 16.5 & 71   & 46.5     
\\
\hline
8 &
cooking recipes      & 0.5 & 7.5 & 2   & 4   & 15   & 52   & 52       
\\
\hline
9 &
airline              & 0.8 & 8   & 2   & 4   & 12   & 55.5 & 53.5     
\\
\hline
10 &
department store         & 0.4 & 4   & 2   & 1   & 4.5  & 38.5 & 54       
\\
\hline \hline
&
median                 & 0.85 & 7  & 2   & 2.75 & 10.25 & 46.75 & 41.75 
\\
\hline 
\end{tabular}
\caption{
\label{results}
Results of the usability study.  A~$=$~task completion rate; B~$=$~median
number of actions; C~$=$~maximum number of pages; D~$=$~median number
of task variables; E~$=$~median command length in words; F~$=$~median
demo time in seconds; G~$=$~median acclimated demo time in seconds.
} 
\end{table}

Table~\ref{results} shows the statistics that we gathered from our
study.  
Column A shows the fraction of attempts in which
Helpa correctly executed the new command.  Despite the current
limitations of Helpa and BRAP, the median success rate over all 10
websites was 85\%.  There were only two kinds of failures. 56\% of the
failures (about 8\% of all attempts) occurred when a web site did
something unexpected that BRAP could not handle.  For example, in the
middle of our study, \url{allrecipes.com} started presenting a new
kind of pop-up ad, which often prevented BRAP from playing a UI script
to completion.  The other 44\% of failures (about 7\% of all attempts)
occurred when a study subject failed to follow the instructions. The
instruction that users failed to follow the most often was the one
pertaining to the limitations of BRAP.  An interesting case
  study here is the department store website \url{jcpenney.com}.
  Subjects had far more trouble with this site than with any other.
  That's because its first page was very simple, with just one search
  box, but its second page had a bewildering array of options for
  narrowing down the search results.  Most subjects excitedly
  attempted to use one or more of these options.  Unfortunately, most
  of the options were rendered by elements that were incompatible with
  BRAP, and many subjects forgot about that restriction.  Overall,
less than 2\% of all attempts failed for reasons unrelated to BRAP.
These results support the working hypothesis stated at the
beginning of Section~\ref{eval}. 

The remaining statistics in Table~\ref{results} are averaged over only
the successfully completed trials.  Column~B shows the median number
of actions per UI script.  This number includes the initial actions
of navigating to the website and waiting for it to load (as in
Figure~\ref{demo}).  Column~C shows the maximum number of page loads
per UI script, again including the initial loading of the website.
Only two pages were used on most websites, because most of the
websites had no BRAP-compatible elements on the second page.  Column~D
shows the median number of variables per UI script.  Column~E shows
the median number of words per command, after tokenization.  We used a
generic English tokenizer, which merely separated words from
punctuation.  

Column~F of Table~\ref{results} shows the median number of seconds
that it took a user to interact with Helpa's learning mode for the
given website.  Time was measured by the wall clock and includes
network delays.  We found that most users struggled with Helpa a bit
until they understood that it won't work unless they follow the
instructions very precisely.  So we also report the median user effort
after acclimation, in Column~G. This measure is the median time per
demo for each website, excluding users for whom that website was the
first or second that they worked on.\footnote{Tables~2 and~1 are both
  sorted on the measure in Column~G.}  Our results show that users can
usually teach Helpa a new task in less than a minute ($p <
0.01$), often much less. Thus, despite its current limitations, Helpa represents a
major advance on the user effort criterion: We are not aware of any
other IA that can learn to predict programs with variables from
natural language commands nearly as quickly.

Appendix~B shows some of the more interesting examples of the
variability of command templates for some of the websites in our
study.

\section{Conclusions}

Virtual assistants (VAs) have become very popular, but not nearly as
popular as they could be.  We conjecture that one of the main reasons
for their slow adoption is that users cannot customize them.  Our
instructible agent (IA) Helpa offers users a way to customize their
VAs, not only in terms of which tasks the VA can perform, but also in
terms of the commands used to trigger those tasks, and the way the
tasks are executed.  To encourage research on this topic, we are
sharing the data set that grew out of our usability study.

Since Helpa succeeded for most users on most websites, we claim that
Helpa can learn many unrelated tasks without its creators'
involvement.  Since Helpa uses no domain-specific knowledge of any
kind, we claim that it has almost complete coverage of tasks that can
be represented by non-branching programs.  We don't know of any other
IA that can learn programs with variables in arbitrary domains without
its creators' involvement.  We also don't know of any other IA that
can be taught new tasks with variables in less than a minute per task.

The work presented here provides a springboard for several directions
of future research.  An obvious direction is to improve Helpa's
components to reduce or remove its limitations.  Another direction is
to develop learning algorithms that can use the available data more
effectively, or learning algorithms for branching programs.  Yet
another direction is to deploy a speech-enabled Helpa on a massive
scale, gather a much larger number of examples, and work towards a
future where Helpa can execute a user's new task correctly without any
training, because it has already learned how to do so from other
users.  

\section*{Appendix A: Instructions Given to Study Subjects}
\subsection*{Introduction}
Thank you for agreeing to participate in our Helpa experiments!  Helpa
is a virtual assistant (VA).  Like other VAs, it can execute verbal
commands through a suitably instrumented agent, such as a smart-phone
or a web browser.  What makes Helpa different from other VAs is that
you can teach it new tasks without any programming.  We are studying
how people use this feature, in order to make it easier to use.  We
are aiming to make the training procedure so fast and intuitive that it
becomes a significant time-saver for developers of VA apps, and
eventually also for end-users.

Our main innovation is the way that Helpa is trained.  To teach Helpa a
new task, you need only give it an example command and demonstrate how
to execute that command.  Helpa can then figure out how to execute new
commands of that type.  With a bit of practice, we have been able to
teach Helpa some new tasks in less than a minute each!

We are starting with relatively simple tasks that involve no loops or
conditionals.  There are many such simple tasks that people perform
often enough to justify automation.  The current set of experiments
will focus on the common example of filling out forms online. 

For example, suppose a user wants to fill out a form on a travel
website, and gives the command ``Find a hotel for 2 nights starting
August 3, 2015.''  A developer who is trying to program a VA to
execute such a command would partition it as follows:
\includegraphics[width=3.25in]{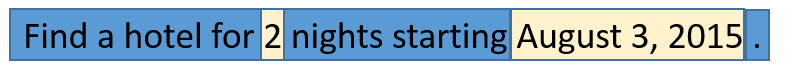}
\newline
Each yellow segment is a variable that needs to be mapped to a field
on the web form.  The blue segments are contextual ``filler'', which a
given user is likely to say in a similar way every time they want this
kind of task done. Without Helpa, a developer would have to specify the
segmentation explicitly.  Helpa can figure out the segmentation and
variable mapping, so that it can be taught by a user who does not have
access to its source code.

We believe the Helpa paradigm can work on any device, but our current
experiments will be done only in a web browser.  Also, eventually,
people will interact with Helpa by speaking to it.  For now, all
interaction is through a text-only ``control'' window.

The success of our experiment depends on your careful adherence to the
instructions.  So please read them very carefully, and tell the
experimenter if there is even a single word that is not perfectly
clear.

\subsection*{Instructions}

To teach Helpa a new task, you must give it an example command, and a
demonstration of how to execute that command in a web browser.  The
experiment will ask you to do so on 10 websites with varying levels of
complexity.  You do no need to use all or even most of every website.
Try to interact with each website the way you imagine a typical
non-technical user might.  Remember that this is a usability study, not
an acid test of Helpa's robustness.  After you teach Helpa a new task,
you will test it on that task, and decide whether it learned the task
correctly.

The current experiments are designed to study whether Helpa can
correctly learn to execute simple commands, as well as to study how
users interact with it.  For this purpose, we have built only a rough
prototype of Helpa, which has many limitations.  In the future, we
plan to improve Helpa by removing most of these limitations.  For now,
keeping in mind the example command segmentation above, please pay
careful attention to the following:
\bi
\item Helpa is currently focused on tasks that involve filling out web
  forms.  Therefore, {\bf every variable in your commands must
    correspond either to a text field or to an option in a pull-down
    menu}.  Your demos can also use check-boxes, radio buttons, and
  push-buttons (like ``Submit'' or ``Search'') but these elements
  cannot represent command variables.  Your demos cannot involve any
  other type of web page elements, such as hyperlinks, tabs, sliders,
  maps, calendars, etc..  Your demos cannot use the Enter key to
  signal form completion.  Also, Helpa cannot handle pop-up menus or
  any other kind of pop-up.  In particular, menus that appear for
  automatic completion of text fields should not be used.
\item {\bf The variable instantiations must appear in the example command
  exactly the same way as they appear in the web form.}  E.g., if the
  web form displays dates like ``August 3, 2014'', the example command
  cannot refer to ``08/03/2014'' or even to ``August 3 2014''.
\item {\bf Ignore default values.}  If your demo needs to use a web form
  element that appears with the correct value already in it, enter
  the value anyway, as if it wasn't there.
\item {\bf The variables in a command cannot be adjacent.}  They must be
  separated by some filler.  E.g., you can't use a command like ``Find
  a 2014 Porsche for rent'' where ``2014'' and ``Porsche'' refer to
  the different fields of a web form.  However, you could rephrase
  such a command as ``Find a Porsche from 2014 for rent'' so that the
  filler ``from'' separates the two variables.
\item {\bf No web form element may be a composite of two or more command
  variables, or vice versa.}  E.g., if there are separate pull-down
  menus for the month and the day of the month, the command cannot
  combine them in the same variable such as ``August 3''.  Or if there
  is a menu option like ``price range from \$100 to \$200'', the command
  cannot have separate variables for the min and the max.
\item (for training only) {\bf No variable value can appear in the command
  more than once}, either as another variable value or as part of the
  contextual filler.  E.g., you cannot use commands like ``Find a hotel
  for 2 nights for 2 people...'' or ``Find a synonym for the word
  find.''
\ei
Also, Helpa is currently a bit slow.  To avoid confusing it, please
pay careful attention to the prompts in the control window, and don't
touch the browser until it finishes loading and the control window
says to go ahead.  This is important every time the browser loads a
new page, which can be triggered unexpectedly in many ways, sometimes
as simple as clicking a radio button.

\subsection*{Frequently Asked Questions}

$\star$ Is there a limit on the number of variables in a command? \newline
No, there is no limit.  However, as previously mentioned, this is a
usability study, not an acid test.  So please don't make your
commands more complicated than they would be for a typical
non-technical user.

\noindent $\star$ Can a demo involve more than one web page? \newline
Yes.  However, remember that you cannot click on hyperlinks, so the
only way to get to another web page during your demo is by clicking a
button such as ``Submit'' or ``Search''.  If there are suitable web
page elements on the next page, then you can continue your demo there.

\vspace{.2in}
You should have a printed copy of these instructions handy during the
experiment, so that you can refer to them whenever you have any doubts.

Feel free to ask the experimenter any questions that you might have.
When you think you understand the instructions well enough to start,
press ENTER in the control window.

\section*{Appendix B: Examples of Command Templates}
Here are some examples of command templates that Helpa
inferred in learning mode during our usability study.  Underscores
represent free variables.

\vspace*{2mm}
thesaurus:
\small
\begin{verbatim}
search for ___ .
dictionary ___
what is a synonym for " ___ " ?
what is another word for ___ ?
search collins for ___
\end{verbatim}
\normalsize

\newpage
recruiting:
\small
\begin{verbatim}
search for ___ as the exact phrase and ___ of work experience
     and a ___ degree in the state of ___ .
i ' m looking to hire a ___ student in ___ with ___ experience
find resumes of people with at least one of ___ 
     and ___ experience in ___
find ___ candidates with experience in " ___ " who worked at ___
find me resumes with the kword ___ and last job title ___ and 
    one job titled ___ with ___ experience with a ___ degree 
    located in ___
find job candidates who did ___ work in ___
find ___ grads in ___
\end{verbatim}
\normalsize

\vspace*{2mm}
investment research:
\small
\begin{verbatim}
search for ___
current stock quote for ___
show ___ performance for 1m period
what is the value of ___ stock
\end{verbatim}
\normalsize

\vspace*{2mm}
scientific database:
\small
\begin{verbatim}
search for ___ in the text field with ___ in the keywords 
    field for publications between the year ___ and ___ 
    sorted by citations .
search for publications by ___ about ___
find papers by  ___ from ___ to ___
find articles by ___ about ___
\end{verbatim}
\normalsize

car rental:
\small
\begin{verbatim}
search for ___ as pick - up with ___ as pick - up date at ___ 
    and ___ as drop - off date at ___ .
find me a car at ___ airport pickup ___ and drop off ___ 
show cars at ___ at ___ on ___ 
i need to rent a car from ___ on ___ at ___ until ___ at ___
\end{verbatim}
\normalsize

\vspace*{2mm}
cooking recipes:
\small
\begin{verbatim}
search for ___ with prep time ___ and meal is ___ and " with 
    these ingredients : " is ___  and ___ for " 
    but not these ingredients : " .
i want to make a ___ with main ingredient ___ with ___
\end{verbatim}
\normalsize

\bibliographystyle{splncs03}
\bibliography{paper}

\end{document}